\pdfoutput=1

\documentclass[11pt]{article}

\usepackage[]{EMNLP2023}

\usepackage{times}
\usepackage{latexsym}

\usepackage{multirow}
\usepackage{adjustbox}
\usepackage{booktabs}

\usepackage[T1]{fontenc}

\usepackage[utf8]{inputenc}

\usepackage{microtype}

\usepackage{inconsolata}

\title{On the Analysis of Cross-Lingual Prompt Tuning for Decoder-based Multilingual Model}

\author{Nohil Park$^1$ \hspace{4mm} Joonsuk Park$^{4,5,6}$ \hspace{4mm} Kang Min Yoo$^{3,4,5,}$\Thanks{\ Corresponding author.} \hspace{4mm} Sungroh Yoon$^{1,2,}\footnotemark[1]$ \\
  $^1$Department of Electrical and Computer Engineering, Seoul National University \\
  $^2$Interdisciplinary Program in Artificial Intelligence, Seoul National University \\
  $^3$Artificial Intelligence Institute, Seoul National University \\
  $^4$NAVER Cloud \hspace{2mm} $^5$NAVER AI Lab \hspace{2mm} $^6$University of Richmond \\
  \{\texttt{pnoil2588}, \texttt{sryoon}\}\texttt{@snu.ac.kr} \hspace{1mm} \texttt{park@joonsuk.org} \hspace{1mm} \texttt{kangmin.yoo@navercorp.com}}

\begin{document}
\maketitle
\begin{abstract}

An exciting advancement in the field of multilingual models is the emergence of autoregressive models with zero- and few-shot capabilities, a phenomenon widely reported in large-scale language models. 
To further improve model adaptation to cross-lingual tasks, another trend is to further fine-tune the language models with either full fine-tuning or parameter-efficient tuning.
However, the interaction between parameter-efficient fine-tuning (PEFT) and cross-lingual tasks in multilingual autoregressive models has yet to be studied.
Specifically, we lack an understanding of the role of linguistic distributions in multilingual models in the effectiveness of token-based prompt tuning.
To address this question, we conduct experiments comparing prompt tuning and fine-tuning on the decoder-based multilingual model, XGLM, with four cross-lingual tasks (XNLI, PAWS-X, POS, NER).
According to our study, prompt tuning achieves on par or better performance over fine-tuning across all languages while updating at most 0.13\% of the model parameters.
Moreover, we empirically show that prompt tuning is more effective in enhancing the performance of low-resource languages than fine-tuning.
Our further analysis shows that the phenomenon is related to the tokenization scheme of the multilingual model.

\end{abstract}

\section{Introduction}\label{sec:introduction}
Recent efforts to use encoder and encoder-decoder models on multilingual tasks have shown that prompt tuning can be more effective than fine-tuning.
At the same time, a strand of research aims to tackle multilingual tasks leveraging multilingual models while avoiding the cost of full fine-tuning
(\citealp{pfeiffer-etal-2020-mad}; \citealp{huang-etal-2022-zero}; \citealp{fu-etal-2022-polyglot}).
In particular, to widely used models such as XLM-R (\citealp{conneau-etal-2020-unsupervised}), variations of prompt tuning have been applied (\citealp{fu-etal-2022-polyglot}; \citealp{huang-etal-2022-zero}; \citealp{zhao-schutze-2021-discrete}). 
This method, when applied as-is, demonstrates better performance on cross-lingual tasks compared to full fine-tuning (\citealp{tu-etal-2022-prompt}).

In this way, previous research has predominantly concentrated on models employing encoder-only (\citealp{conneau-etal-2020-unsupervised}) or encoder-decoder (\citealp{xue-etal-2021-mt5}) architectures.
Thus, there remains a necessity to explore the effectiveness of the prompt tuning for cross-lingual transfer learning in the decoder-based multilingual models, such as XGLM (\citealp{lin-etal-2022-shot}) and BLOOM (\citealp{scao2022bloom}). 
This will not only fill the gap in the study of prompt tuning on decoder-only models, but also has the potential to allow more efficient use of such models.

In this paper, we compare the effectiveness of prompt tuning against that of fine-tuning on a decoder-only multilingual model, XGLM. 
The goal is analogous to that of an existing study on an encoder-based model, XLM-R (\citealp{tu-etal-2022-prompt}).
We conduct experiments utilizing P-tuning v2 (\citealp{liu-etal-2022-p}), which entails adding fixed-length continuous prompts in front of the input sequences and all layers of XGLM. 
During this process, we update only the prompt embeddings and the classifier heads while keeping the remaining parameters frozen.
In the analysis, we verify our hypothesis on the influence of language imbalance and excessively fine-grained tokenization on prompt tuning for low-resource languages.

We find that models trained with prompt tuning perform comparable to or even better than models trained with fine-tuning across the four cross-lingual tasks, namely XNLI (\citealp{conneau-etal-2018-xnli}), PAWS-X (\citealp{yang-etal-2019-paws}), NER (\citealp{rahimi-etal-2019-massively}), and POS (\citealp{liang-etal-2020-xglue}).
In the XNLI task, for example, models trained with prompt tuning achieve accuracy improvements of up to 2.9\% over full fine-tuned models, while updating only 0.08\%-0.13\% of the parameters. 
Also, we observe a correlation between the language distribution and the cross-lingual performance. 
In particular, prompt tuning yields greater performance gains compared to fine-tuning for low-resource languages.

This paper presents the following contributions.
\begin{itemize}
    \item As far as we know, this paper represents the first study on the effectiveness of the \textit{parameter-efficient} prompt tuning for the decoder-only multilingual model XGLM.
    \item We observe that prompt tuning shows similar or even better performance compared to fine-tuning across the four cross-lingual tasks. Furthermore, prompt tuning shows its effectiveness in enhancing performance gains for low-resource languages.
    \item We identify distinct patterns of prompt tuning and analyze that language distribution and excessive tokenization on low-resource languages are related to this phenomenon.
\end{itemize}
\section{Approach}\label{sec:approach}

\subsection{Model}
XGLM adopts the identical Transformer decoder structure as GPT-3 (\citealp{NEURIPS2020_1457c0d6}), but it incorporates a larger set of 250K joint vocabularies in its embedding parameters to encompass 500B-token multilingual corpora consisting of 30 languages.
Given these characteristics, the XGLM models are structured to align with the layer count and hidden dimension size of GPT-3 models, resulting in slightly larger model sizes due to the inclusion of additional embedding parameters.

Unlike conventional multilingual models such as mBERT (\citealp{pires-etal-2019-multilingual}), XLM-R (\citealp{conneau-etal-2020-unsupervised}), and mT5 (\citealp{xue-etal-2021-mt5}), XGLM employs a decoder-only architecture.
While causal language modeling of the decoder-only model is advantageous for generation tasks, it is known to perform as well as masked language modeling models in natural language understanding tasks (\citealp{liu2021gpt}).

In this paper, we utilize the XGLM-564M, the smallest among the publicly available pre-trained XGLM models provided on \texttt{HuggingFace}\footnote[1]{\url{https://huggingface.co/facebook/xglm-564M}}, as the backbone for our models.

\subsection{Prompt Tuning}
Unlike the initial approach of appending continuous tokens only at the beginning of the input sequence (\citealp{liu2021gpt}), P-tuning v2 goes a step further by adding such tokens in front of each model layer.
This expands the tunable parameters to some extent but enhances the representation learning capabilities for each task.

In this paper, we adopt the fixed-length continuous prompts attached to both the input sequence and the front of each layer of the XGLM model, following the settings of P-tuning v2.
For each task during prompt tuning, we update the classification head attached to the model along with the continuous prompts, while freezing the rest of the pre-trained model weights.

Previous work on prompt tuning reparameterizes the continuous prompts to improve performance, instead of directly optimizing them (\citealp{li-liang-2021-prefix}; \citealp{liu2021gpt}).
We also incorporate a two-layer MLP prompt encoder as the reparameterization trick in the experiments.

\subsection{Cross-Lingual Transfer Tasks}
We experiment with four cross-lingual tasks: XNLI (\citealp{conneau-etal-2018-xnli}), PAWS-X (\citealp{yang-etal-2019-paws}), NER (\citealp{rahimi-etal-2019-massively}), and POS (\citealp{liang-etal-2020-xglue}).
These tasks can be grouped into two categories: sequence-pair classification and structured prediction tasks, with two datasets each in sequential order.
We use the datasets provided by \texttt{HuggingFace}\footnote[2]{\url{https://huggingface.co/datasets}}.

\section{Experiment}\label{sec:experiment}

\subsection{Experimental Settings}\label{subsec:experimental-settings}
Referring to the previous studies (\citealp{li-liang-2021-prefix}; \citealp{liu2021gpt}; \citealp{liu-etal-2022-p}), we conduct the experiments on various sets of hyperparameters to find the best-performing model in each task and tuning method, as shown in Table \ref{hyperparameters}.
Furthermore, we conduct multi-seed training to introduce randomness to weight initialization on the continuous prompts and the classification heads, and the ordering of training data (\citealp{chen-etal-2022-revisiting}).
We apply early stopping with respect to the evaluation loss of the validation set to obtain the best checkpoint for each task.
For each setup, we use the average and standard deviation of results from the three models as the final performance metrics.

\begin{table}[ht]
\centering
\begin{adjustbox}{width=1\linewidth}
\begin{tabular}{c|c}
\hline
    \textbf{Hyperparamters} & \textbf{Search Space} \\
\hline
    \textbf{Learning Rate} & {5e-5, 1e-4, 5e-4, 1e-3, 5e-3, 1e-2, 5e-2} \\
    \textbf{Prompt Length} & {8, 10, 15, 20, 25, 30, 35, 40, 45, 50} \\
    \textbf{Batch Size} & {64, 128} \\
\hline
\end{tabular}
\end{adjustbox}
\caption{\label{hyperparameters}
Hyperparameter search space for each model.
}
\end{table}

\begin{table*}[!t]
\centering
\begin{adjustbox}{width=1\linewidth}
\begin{tabular}{@{}ccccccccc@{}}
\hline
    \multirow{2}{*}{\textbf{Task}} & \multicolumn{4}{c}{\textbf{Sentence-pair Classification}} & \multicolumn{4}{c}{\textbf{Structured Prediction}} \\ \cmidrule(l){2-9} 
     & \multicolumn{2}{c}{\textbf{XNLI}} & \multicolumn{2}{c}{\textbf{PAWS-X}} & \multicolumn{2}{c}{\textbf{NER}} & \multicolumn{2}{c}{\textbf{POS}} \\
\hline
    Metric & \multicolumn{2}{c}{Acc. $\uparrow$} & \multicolumn{2}{c}{Acc. $\uparrow$} & \multicolumn{2}{c}{F1 $\uparrow$} & \multicolumn{2}{c}{Acc. $\uparrow$} \\
\hline
    \textbf{Language} & \textbf{FT} & \textbf{PT} & \textbf{FT} & \textbf{PT} & \textbf{FT} & \textbf{PT} & \textbf{FT} & \textbf{PT} \\
\hline
    \textbf{en} & 81.92$\pm$0.92 & 82.15$\pm$0.92 & 92.38$\pm$0.32 & 93.42$\pm$0.51 & 74.29$\pm$0.55 & 74.47$\pm$0.36 & 92.97$\pm$0.06 & 92.79$\pm$0.04 \\
    \textbf{ru} & 69.55$\pm$2.04 & 73.24$\pm$0.29 & - & - & - & - & 72.36$\pm$2.20 & 75.25$\pm$0.57 \\
    \textbf{zh} & 69.11$\pm$0.87 & 70.72$\pm$0.97 & 76.05$\pm$0.68 & 76.68$\pm$1.08 & - & - & 30.48$\pm$2.48 & 27.33$\pm$1.88 \\
    \textbf{de} & 74.26$\pm$0.99 & 74.75$\pm$0.17 & 83.98$\pm$0.69 & 83.87$\pm$0.46 & 32.61$\pm$0.34 & 39.59$\pm$0.87 & 77.55$\pm$1.78 & 77.45$\pm$0.57 \\
    \textbf{es} & 75.65$\pm$0.51 & 76.26$\pm$0.07 & 85.62$\pm$0.58 & 87.03$\pm$0.63 & 54.19$\pm$0.71 & 56.19$\pm$0.18 & 77.89$\pm$2.07 & 78.01$\pm$0.77 \\
    \textbf{fr} & 75.87$\pm$1.18 & 75.90$\pm$0.24 & 87.05$\pm$0.71 & 86.32$\pm$1.18 & - & - & 75.28$\pm$0.76 & 74.34$\pm$1.05 \\
    \textbf{ja} & - & - & 72.23$\pm$1.58 & 71.13$\pm$2.52 & - & - & - & - \\
    \textbf{it} & - & - & - & - & - & - & 71.97$\pm$1.57 & 72.29$\pm$0.69 \\
    \textbf{el} & 71.38$\pm$2.23 & 74.54$\pm$0.16 & - & - & - & - & 69.95$\pm$3.16 & 79.41$\pm$3.09 \\
    \textbf{ko} & - & - & 71.15$\pm$1.60 & 70.38$\pm$0.98 & - & - & - & - \\
    \textbf{pl} & - & - & - & - & - & - & 29.29$\pm$3.98 & 28.83$\pm$3.89 \\
    \textbf{nl} & - & - & - & - & 32.84$\pm$0.13 & 34.92$\pm$0.51 & 43.02$\pm$5.10 & 44.17$\pm$4.71 \\
    \textbf{tr} & 64.51$\pm$3.44 & 69.18$\pm$1.27 & - & - & - & - & 57.31$\pm$4.47 & 61.72$\pm$0.91 \\
    \textbf{ar} & 64.59$\pm$2.51 & 69.58$\pm$0.28 & - & - & - & - & 59.88$\pm$2.50 & 56.76$\pm$1.37 \\
    \textbf{vi} & 70.41$\pm$1.36 & 71.59$\pm$0.46 & - & - & - & - & 57.82$\pm$3.07 & 57.19$\pm$0.92 \\
    \textbf{th} & 59.35$\pm$5.40 & 67.50$\pm$0.96 & - & - & - & - & 44.62$\pm$3.38 & 44.85$\pm$0.93 \\
    \textbf{bg} & 73.99$\pm$1.55 & 76.10$\pm$0.37 & - & - & - & - & 77.62$\pm$1.58 & 80.41$\pm$0.96 \\
    \textbf{hi} & 64.60$\pm$1.57 & 67.28$\pm$1.01 & - & - & - & - & 51.49$\pm$5.90 & 55.39$\pm$0.85 \\
    \textbf{ur} & 56.87$\pm$6.08 & 62.83$\pm$1.05 & - & - & - & - & 49.65$\pm$4.19 & 53.30$\pm$0.56 \\ 
    \textbf{sw} & 61.03$\pm$4.46 & 64.93$\pm$0.76 & - & - & - & - & - & - \\
\hline
    \textbf{Avg.} & 68.87$\pm$1.02 & 71.77$\pm$0.36 & 81.21$\pm$0.31 & 81.26$\pm$1.01 & 48.48$\pm$0.16 & 51.29$\pm$0.29 & 61.13$\pm$2.46 & 62.32$\pm$0.65 \\
\hline
\end{tabular}
\end{adjustbox}
\caption{\label{main-results}
Cross-lingual transfer evaluation results of XGLM-564M models with prompt tuning (PT) and fine-tuning (FT) on XNLI, PAWS-X, NER, and POS tasks.
Each model is trained on the English training set and evaluated on the test set of all target languages to measure performance.
The final performance of the models is presented as the average and standard deviation of the results obtained from three models trained with three different seeds.
Languages are sorted in descending order based on their proportion in CC100-XL (\citealp{lin-etal-2022-shot}), XGLM's pre-training dataset.
The performance of languages not covered in the test sets is indicated with a hyphen (-).
}
\end{table*}

\subsection{Results}\label{subsec:results}
The evaluation results of the XGLM-564M model tuned with prompt tuning and fine-tuning, respectively, can be found in Table \ref{main-results}.
On the four cross-lingual tasks, prompt tuning generally performs on par with or outperforms fine-tuning.
Even considering the performance variance, prompt tuning exhibits better performance than fine-tuning especially in the XNLI and NER tasks.

We also note that high-resource languages, such as English, Russian, Spanish, and French, consistently exhibit relatively better performance across all tasks, and vice versa.
Among many possible causes, we speculate the language imbalance in the pre-training dataset of the XGLM model.

\citet{lin-etal-2022-shot} also point out that low-resource languages exhibit a relatively lower overlap of subwords with other languages, and the tokenizer trained on imbalanced corpora further breaks down words into smaller subword units, leading to a higher level of granularity.
We share a similar perspective that the fine-grained tokenization may restrict the model's capacity to capture more comprehensive contextual information, thereby potentially contributing to the performance decline in low-resource languages as shown in Table \ref{main-results}.

\subsection{Analysis}\label{subsec:analysis}
By comparing the performance differences between the two models for each language in Table \ref{main-results}, we can evaluate the performance gains obtained through each tuning technique.
For the XNLI task, we can observe that languages such as English, German, Spanish, and French have performance differences of less than 1\% in terms of performance gains. 
Languages like Turkish, Arabic, Thai, and Urdu, on the other hand, show larger differences in performance gains, ranging from 4.67\% to as high as 8.15\%.
In summary, excluding a few exceptional languages such as Chinese and Vietnamese, we can conclude that low-resource languages generally exhibit larger performance improvements through prompt tuning compared to high-resource languages, surpassing the magnitude of performance gains achieved through fine-tuning.

In the case of low-resource languages, where word granularity is high as mentioned in \citet{lin-etal-2022-shot}, although it may be disadvantageous for capturing high-level contextual information, it can be advantageous for learning language identification information instead.
In other words, we presume the finely tokenized subwords may act as a form of discrete prompts (\citealp{liu2021gpt}; \citealp{han2021ptr}), leading to the observation that prompt tuning is more effective for low-resource languages compared to fine-tuning.
\citet{liu2021gpt} report that using discrete tokens as anchor prompts along with continuous prompts improve the model performance on various NLU tasks.
In this paper, although such anchor prompts are not used explicitly, we speculate that the small subword units of low-resource languages might implicitly aid the performance gains of prompt tuning.
\section{Related Work}\label{sec:related_work}

\subsection{Multilingual Models}
Multilingual models are based on the widely used Transformers architecture (\citealp{NIPS2017_3f5ee243}) and the language modeling objectives commonly employed in monolingual models (\citealp{devlin-etal-2019-bert}; \citealp{NEURIPS2020_1457c0d6}; \citealp{DBLP:journals/jmlr/RaffelSRLNMZLL20}). 

Significantly improved performance in these models has been achieved by combining various techniques such as tokenizers designed to accommodate numerous languages (\citealp{sennrich-etal-2016-neural}; \citealp{kudo-2018-subword}; \citealp{song-etal-2021-fast}), considerably larger vocabularies (\citealp{pires-etal-2019-multilingual}; \citealp{conneau-etal-2020-unsupervised}; \citealp{lin-etal-2022-shot}), special tokens for language indication and training objectives that are helpful for learning language alignment information (\citealp{NEURIPS2019_c04c19c2}).
Similar to the recent advancement in the large-scale language models (\citealp{NEURIPS2020_1457c0d6}; \citealp{smith2022using}; \citealp{chowdhery2022palm}), multilingual models are also maximizing the zero-shot and few-shot performance on various downstream tasks through larger parameters and a greater amount of multilingual training corpora (\citealp{lin-etal-2022-shot}; \citealp{scao2022bloom}).

\subsection{Prompt Tuning}
Prompt tuning (\citealp{lester-etal-2021-power}) is a parameter-efficient tuning technique that aims to achieve performance similar to fine-tuning, while updating only a small portion of a large language model.
It primarily involves appending fixed-length tunable continuous prompts called \textit{prefix} to the input sentence, while freezing the remaining parameters, to maximize computational efficiency (\citealp{li-liang-2021-prefix}; \citealp{liu2021gpt}).
\citet{liu-etal-2022-p} proposed P-tuning v2 with additional trainable parameters attached to every model layer to bring performance boosting across various tasks.

Prompt tuning was found to be effective in the encoder-based multilingual model and the language-independent decision boundary was stated as the primary factor that contributes to the superiority of it (\citealp{tu-etal-2022-prompt}).
Although \citet{winata-etal-2022-cross} reports \textit{prompt tuning} performance on the XGLM model, it does not fall under the \textit{parameter-efficient tuning} as both the discrete prompt templates attached to the input sentence and the remaining pre-trained weights are tuned.
Additionally, the evaluation dataset (\citealp{winata-etal-2023-nusax}) is specialized in the Indonesian language family, while the cross-lingual datasets we employed cover a broader range of language groups.
\section{Conclusion}\label{sec:conclusion}

In this paper, we reconfirm the performance of prompt tuning observed in the encoder-based multilingual models in the context of the decoder-based generative multilingual model, XGLM.
We also demonstrate that performance tends to improve overall as the language becomes dominant in the training set, with prompt tuning particularly yielding significant performance gains compared to fine-tuning in low-resource languages.
We suspect that the fundamental cause of this phenomenon lies in the language imbalance which leads to fine-grained tokenization of low-resource languages.
As a result, in low-resource languages, finely segmented subword units function similarly to discrete prompts and lead to the effectiveness of prompt tuning.

In future work, we intend to expand our experiments to cross-lingual natural language generation tasks (\citealp{liang-etal-2020-xglue}) to observe the potential strengths of the generative language model.
Furthermore, investigating whether prompt tuning can maintain competitiveness over fine-tuning when evaluated on other large-scale decoder-only models such as BLOOM (\citealp{scao2022bloom}) is also an interesting topic.
We also aim to study whether other PEFT methods such as LoRA (\citealp{hu2022lora}) or Adapters (\citealp{pmlr-v97-houlsby19a}) work well on the decoder-based multilingual models.

\section*{Limitations}
Our experiments are based on the pre-trained XGLM-564M model, the smallest among the publicly released versions, due to limited computation resources with a single A40 GPU.
Previous studies claim that as the model size increases, prompt tuning becomes even more parameter-efficient, maximizing its effectiveness (\citealp{lester-etal-2021-power}; \citealp{kaplan2020scaling}).
Therefore, expanding our experiments to publicly available larger models, such as XGLM-7.5B, in order to demonstrate competitiveness can become a promising future work.

In this paper, we achieved performance comparable to fine-tuning across various cross-lingual tasks, despite the most basic form of prompt tuning method.
We look forward to expand our experiments using prompt tuning techniques specifically designed for multilingual models (\citealp{fu-etal-2022-polyglot}; \citealp{vu-etal-2022-overcoming}; \citealp{huang-etal-2022-zero}), as well as exploring other branches of parameter-efficient tuning techniques (\citealp{hu2022lora}; \citealp{pmlr-v97-houlsby19a}; \citealp{pfeiffer-etal-2020-mad}).

\section*{Ethics Statement}
By adopting the \textit{parameter-efficient} prompt tuning, we expect increased energy efficiency and reduced carbon footprint by utilizing fewer GPU resources.
Due to the inherent language imbalance in the training corpus, there is a performance degradation in low-resource languages. 
We show that prompt tuning can compensate for this performance decline by leveraging the interaction with tokens specific to low-resource languages through our analysis.


\bibliography{anthology,custom}
\bibliographystyle{acl_natbib}

\appendix

\section{Model Details}\label{sec:appendix-model-details}

\begin{table}[h]
\centering
\begin{adjustbox}{width=1\linewidth}
\begin{tabular}{@{}ccccc@{}}
\toprule
\textbf{Task} & \textbf{Model} & \textbf{Prompt Length} & \textbf{Learning Rate} & \textbf{Batch Size} \\ \midrule
\multirow{2}{*}{XNLI} & FT & - & 1e-5 & 64 \\
 & PT & 15 & 1e-2 & 128 \\
\multirow{2}{*}{PAWS-X} & FT & - & 5e-5 & 128 \\
 & PT & 15 & 5e-3 & 128 \\
\multirow{2}{*}{NER} & FT & - & 5e-5 & 64 \\
 & PT & 15 & 1e-2 & 128 \\
\multirow{2}{*}{POS} & FT & - & 5e-5 & 64 \\
 & PT & 10 & 1e-3 & 64 \\ \bottomrule
\end{tabular}
\end{adjustbox}
\caption{\label{best-hyperparameters}
Hyperparameters for the best checkpoints of XGLM-564M trained with prompt tuning (PT) and fine-tuning (FT) on each task.
}
\end{table}

Table \ref{best-hyperparameters} is a summary of the setups that achieve the best performance across the four cross-lingual tasks through multi-seed experiments.
The hidden dimension size of the MLP prompt encoder is fixed at 512, thus the proportion of tunable parameters varies depending on the length of the continuous prompts.
Out of the total 564M parameters in the model, approximately 0.087\%, 0.1304\%, and 0.1738\% are tunable when the prompt length is 10, 15, and 20, respectively.
All experiments are done on a single A40 GPU.

\section{Subword Fertility}\label{sec:subword-fertility}

\begin{table}[h]
\centering
\begin{adjustbox}{width=1\linewidth}
\begin{tabular}{@{}ccccc@{}}
\toprule
\textbf{Language}        & \textbf{XNLI} & \textbf{PAWS-X} & \textbf{NER} & \textbf{POS} \\ 
\midrule
\textbf{English (en)}    & 1.24          & 1.31            & 1.35         & 1.26         \\
\textbf{Russian (ru)}    & 1.51          & -               & -            & 1.71         \\
\textbf{Chinese (zh)}    & 2.09          & 14.69           & -            & 2.13         \\
\textbf{German (de)}     & 1.41          & 1.59            & 1.55         & 1.4          \\
\textbf{Spanish (es)}    & 1.25          & 1.41            & 1.34         & 1.31         \\
\textbf{French (fr)}     & 1.31          & 1.51            & -            & 1.4          \\
\textbf{Japanese (ja)}   & -             & 16.46           & -            & -            \\
\textbf{Italian (it)}    & -             & -               & -            & 1.35         \\
\textbf{Greek (el)}      & 1.57          & -               & -            & 1.62         \\
\textbf{Korean (ko)}     & -             & 2.33            & -            & -            \\
\textbf{Polish (pl)}     & -             & -               & -            & 1.71         \\
\textbf{Dutch (nl)}      & -             & -               & 1.42         & 1.46         \\
\textbf{Turkish (tr)}    & 1.5           & -               & -            & 1.73         \\
\textbf{Arabian (ar)}    & 1.62          & -               & -            & 1.45         \\
\textbf{Vietnamese (vi)} & 1.36          & -               & -            & 1.55         \\
\textbf{Thai (th)}       & 1.61          & -               & -            & 1.64         \\
\textbf{Bulgarian (bg)}  & 1.5           & -               & -            & 1.55         \\
\textbf{Hindi (hi)}      & 1.46          & -               & -            & 1.38         \\
\textbf{Urdu (ur)}       & 1.21          & -               & -            & 1.41         \\
\textbf{Swahili (sw)}    & 1.43          & -               & -            & -            \\
\bottomrule
\end{tabular}
\end{adjustbox}
\caption{\label{table:subword-fertility}
Language-specific subword fertility. Languages are sorted in descending order based on their proportion in CC100-XL, the pre-training dataset of XGLM models).
}
\end{table}

To support the hypothesis in Section \ref{subsec:analysis}, we include language-specific subword fertility results (average number of subwords per word) as shown in Table \ref{table:subword-fertility}. Since the pre-training dataset for the XGLM model, CC100-XL, is proprietary, direct calculation of subword fertility for the languages comprising the pre-trained model's vocabulary is challenging. Instead, we aim to use tokenization results obtained with the pre-trained XGLM tokenizer on samples from the four cross-lingual transfer datasets employed in our experiments for assessing subword fertility.

For the XNLI task, low-resource languages (Turkish, Arabic, Thai, and Urdu) with significant performance gains have an average fertility of 1.485, whereas high-resource languages (English, German, Spanish, and French) with smaller performance gains have an average fertility of 1.3025.

For the POS task, low-resource languages (Turkish, Hindi, and Urdu) with notable performance gains have an average fertility of 1.5067, while high-resource languages (Spanish and Italian) with modest performance gains exhibit an average fertility of 1.33.

It's worth noting that Chinese and Japanese have higher fertility compared to other languages, mainly because they often convey meaning at the character level, resulting in fine-grained tokenization regardless of their proportion in the pre-training dataset (\citealp{lin-etal-2022-shot}). Excluding these exceptional cases, we can infer a correlation between the extent of tokenization, represented by fertility, and the magnitude of performance gains achieved through prompt tuning.

\end{document}